# Expressway visibility estimation based on image entropy and piecewise stationary time series analysis

XIAOGANG CHENG[1, 2], (Member, IEEE), GUOQING LIU[3], Anders Hedman[2], KUN WANG[4], (Senior Member, IEEE), HAIBO LI[1, 2]

[1]College of Telecommunications and Information Engineering, Nanjing University of Posts and Telecommunications, Nanjing, 210003, China
[2]School of Electrical Engineering and Computer Science, Royal Institute of Technology, Stockholm, 10044, Sweden
[3]School of Physical and Mathematical Sciences, Nanjing Tech University, Nanjing, 211800, China
[4]College of IoT, Nanjing University of Posts and Telecommunications, Nanjing, 210003, China

Corresponding author: Xiaogang Cheng (xiacheng@kth.se, chengxg@njupt.edu.cn)

This study was supported by the National Natural Science Foundation of China (No. 61401236), the Jiangsu Postdoctoral Science Foundation (No. 1601039B), the Key Research Project of Jiangsu Science and Technology Department (No. BE2016001-3), the Ministry of Education-China Mobile Research Fund (No. MCM20150504), and the Scientific Research Foundation of NUPT (No. NY214005).

**ABSTRACT** Vision-based methods for visibility estimation can play a critical role in reducing traffic accidents caused by fog and haze. To overcome the disadvantages of current visibility estimation methods, we present a novel data-driven approach based on Gaussian image entropy and piecewise stationary time series analysis (SPEV). This is the first time that Gaussian image entropy is used for estimating atmospheric visibility. To lessen the impact of landscape and sunshine illuminance on visibility estimation, we used region of interest (ROI) analysis and took into account relative ratios of image entropy, to improve estimation accuracy. We assume fog and haze cause blurred images and that fog and haze can be considered as a piecewise stationary signal. We used piecewise stationary time series analysis to construct the piecewise causal relationship between image entropy and visibility. To obtain a real-world visibility measure during fog and haze, a subjective assessment was established through a study with 36 subjects who performed visibility observations. Finally, a total of two million videos were used for training the SPEV model and validate its effectiveness. The videos were collected from the constantly foggy and hazy Tongqi expressway in Jiangsu, China. The contrast model of visibility estimation was used for algorithm performance comparison, and the validation results of the SPEV model were encouraging as 99.14% of the relative errors were less than 10%.

## I. INTRODUCTION

Fog and haze, especially dumpling fog, can cause serious traffic accidents [1]. To overcome drawbacks of existing optics-based visibility estimation methods, including limitations of sampling volume and high costs, vision-based visibility estimation has become popular. Based on the Koschmieder formula [2-3], many vision-based methods have been presented, e.g., contrast models [4-8], luminance curve models [9-12], road sign models [13-17], and regression model [18-20]. For these approaches, the extinction coefficient in the Koschmieder formula is extracted, and the atmospheric visibility is estimated.

From a practical application standpoint, while vision-based methods are interesting, they have some drawbacks: (1) The existing methods are not verified by big data from real-world conditions and are typically model-driven. The real world is complicated and the model-driven approach is difficult to adapt to its complexity. (2) Lack of global feature variables to describe the fog and haze. Most of the current visibility estimation methods adopt local feature variables. Some of the current methods extract the local features by detecting objects in the foggy and hazy images, but valuable information in foggy and hazy images may be lost in the process. Furthermore, they are scene-dependent, if the scene is changed, the algorithm should also be re-constructed. (3) The luminance curves are not stable and typically contaminated by noise. Luminance curves are used in many current vision-based methods (luminance models). These methods rely on the incorporation of a black object with constant luminance as part of a luminance model. However, it is typically difficult to find a suitable black object and researchers must make compromises, *e.g.*, road lane lines are often used as black objects. When luminance curves contain noise, visibility estimation errors increase. Considering these drawbacks, we sought an alternative simple and reliable approach for estimating visibility during foggy and hazy expressway conditions.

Image entropy analysis has been used in others areas of image processing with clear benefits [21-25]. Image entropy can be seen as a variable for extracting global features that can then be used for fog and haze visibility estimation. With this idea as a starting point, we explored a novel fog and haze visibility estimation algorithm based on Gaussian image entropy and piecewise stationary time series analysis to overcome the drawbacks of current vision-based approaches that we will refer to here as SPEV algorithm. Our contributions in developing and utilizing SPEV are as follows:

(1) It is the first time that Gaussian image entropy is used for visibility estimation. The Gaussian entropy is a global feature variable, and was combined with piecewise stationary time series analysis.

(2) To overcome effects of different road landscapes and sunshine luminance, expressway pavement was defined as the region of interest (ROI) for Gaussian entropy estimation, and relative ratios of Gaussian entropy were utilized.

(3) Unlike current model-driven visibility estimation methods, the method proposed here is a data-driven approach. A big real-world dataset was collected from the Tongqi expressway, Jiangsu, China, to train our data-driven model.

(4) Our approach is grounded in real-world subjective assessments of fog and haze visibility. For the first time, realistic expressway scenes and big video data were used for subjective assessment. In its practical application, the estimated value of fog and haze visibility will be used by car drivers or staff of expressway management center for accident prevention. To validate our approach, 36 subjects subjectively assessed visibility according to the atmospheric visibility definition offered by the international commission on illumination (CIE) [26] and we then compared the results of with those of SPEV.

The rest of this paper is organized as follows. Section II reviews the CIE visibility definition and current model-driven methods. In Section III, SPEV is described in detail along with the subjective assessments. The big data validation results are then described in Section IV. Discussion and conclusions are given in Section V and Section VI, respectively.

## II. RELATED WORK

According to the atmospheric visibility definition offered by the CIE [26], visibility is the greatest distance at which a black object of suitable dimensions can be recognized by day against the horizon sky. The definition is based on the Koschmieder law [2, 3]:

$$L = L_0 e^{-kd} + L_f (1 - e^{-kd}) \quad (1)$$

where $L$ denotes observed luminance of the black object, $L_0$ denotes real luminance of the black object itself and $L_f$ denotes luminance of the sky background. $k$ is the extinction coefficient and $d$ is the distance between the observer and the object. Let $\varepsilon$ denote the visual contrast threshold and $\varepsilon = (L - L_f)/(L_0 - L_f)$ [27]. (1) can then be expressed as $\varepsilon = e^{-kd}$ and we have:

$$Vis = -\frac{\ln(\varepsilon)}{k} \quad (3)$$

The CIE defined the value of $\varepsilon$ as 0.05. The atmospheric visibility, $Vis$ is then $2.99/k$. The Koschmieder law laid the foundation of atmospheric visibility estimation [2, 3, 9, 27]. Many vision-based visibility estimation methods based on the Koschmieder law, have been proposed in recent years.

Some traditional vision-based methods are contrast-based [4-8]. A rapidly adapting lateral position handler system was proposed [4] for forward-looking image recognition. Visibility was estimated by measuring attenuation of contrast between consistent road features at various distances ahead of the vehicle. The contrast of different gray-level areas is related to the corresponding gradient's value [5]. A good wavelet function is was defined based on this. Then a B-splines wavelet transform could be combined with the image contrast to estimate visibility. A depth map of the vehicle environment was constructed in [6]. Then the most distant object on the road surface with a contrast above 5% was combined with the depth map and used to estimate the visibility. This method is based on the definition of the atmospheric visibility distance proposed by CIE. Distribution of contrasts in the fog scene was calculated in [7], and a probabilistic model-based approach of visibility estimation was proposed and validated. Image contrast was combined with linear regression analysis in [8] and a semi-supervised, learning framework was constructed to estimate fog visibility.

The luminance-curve model has become popular during the recent decade [9-12]. Based on the Koschmieder law, a strict mathematical derivation of a visibility-estimation formula was presented in [9]. The inflection point of the luminance curves was used to denote the visual critical point in the road. The formula proposed in [9] can also used for the subjective assessment of visibility. Based on [9], many proposed various alternative methods for finding the inflection point [10-12] have been explored.

Road signs are often used to estimate atmospheric visibility [13-16]. The road signs include pavement [13], roadside signs [14-15], road lane lines [16], *etc*. The methods they use are Sobel operator and Hough line detection [13],

Levenberg-Marquardt [14], Gaussian Mixture Model [15] and dark channel prior [16], respectively.

Sometimes two of the methods above are combined. For example, based on [9] luminance curves are combined with road signs to estimate visibility in [17]. A static calibration method was presented and photo-metrically simulated pictures were used to quantitatively validate the algorithm.

Some researchers extract features from fog images and then apply regression functions to estimate visibility, e.g., in [18]. In [19] a nonlinear relationship function between image position and visibility based on chromatic analysis was constructed to estimate visual range. One visibility estimation method [20] related to entropy utilizes histogram entropy minimization method. The region was selected first, then the depth-map was obtained from a 3D model of the scene (or from a stereo reconstruction). The minimum entropy was used in image restoration and the extinction coefficient was obtained. A suitable region and a robust restoration algorithm are required in [20], but can be difficult to construct.

The drawbacks of the visibility estimation methods above [4-20] were discussed in Section I. Let us now turn to the aim of this study, the construction of our approach to overcome these drawbacks.

## III. Approach
### A. Piecewise Stationary SPEV algorithm

Information entropy denotes the order of a system [28]. The more stable the system is, the lower the information entropy and, vice versa. Image entropy is an application of information entropy theory in image processing. It inherits characteristics of information entropy and the image sequence is considered as an uncertainty source. Image entropy denotes the chaotic degree of the image information.

Let $f(x, y)$ denote the image intensity, the image intensity entropy can then be expressed as [28]:

$$H(x,y) = -\sum_{i=1}^{N}\sum_{j=1}^{M} p\left[x_i, y_j\right] \log p\left[x_i, y_j\right] \quad (1)$$

$$p\left[x_i, y_j\right] = \frac{\left|f\left(x_i, y_j\right)\right|}{\sum_{i=1}^{N}\sum_{j=1}^{M}\left\|f\left(x_i, y_j\right)\right\|_2^2} \quad (2)$$

where $p[x_i, y_i]$ denotes the probability density function of $f(x_i, y_i)$ in the whole image, and $p[x_i, y_i] \leq 1$, and the image size is $N \times M$. In this paper, the images used in formulas (1), (2) are foggy- and hazy-day images or clear-day images.

We assume the foggy and hazy day images are blurred, and the clear day images are clear. The fog and haze can then be considered as noise, *e.g.*, as white or salt and pepper noise. We can then think of the blurred images as superposition of fog and/or haze on clear images.

We can now compute the Gaussian image entropy based on (1) and (2). For simplicity, hereafter referred to as image entropy. On clear days, the image entropy of surveillance videos is small, and the system is orderly. Conversely, on foggy and hazy days, the image entropy of surveillance videos is large, and the system is chaotic.

From a practical standpoint, two real-world factors must be taken into account, (1) the landscapes of the expressway surveillance points are different, and (2) the illumination of each surveillance point varies throughout the day. We deployed region of interest (ROI) extraction and relative-ratio, image-entropy analysis to overcome visibility estimation challenges associated with these two factors.

We assumed that the material of the expressway pavement was evenly distributed. Based on the visibility definition, the pavement area (captured by surveillance cameras) was then used both as the black object and ROI.

When the foggy- and hazy-day videos had been captured, the clear-day videos were captured at the same surveillance points at the same time of day (noon) and for the same duration (fifty minutes). We computed image entropy for all clear-day images. Then mean image entropy values were calculated by removing all singular values to obtain the clear-day image entropy. The relative ratio of image entropy could then then be expressed as:

$$H_r = 10 * \frac{H_{fog}(x,y)}{H_{clear}(x,y)} \quad (3)$$

where $H_{sunny}(x, y)$ denotes clear-day image entropy, and $H_{fog}(x, y)$ denotes foggy- and hazy-day images.

To established a robust and suitable nonlinear relationship between image entropy and real-world visibility, piecewise stationary time series analysis was adopted. Assume that the fog/haze is a stationary signal during different time durations. The fog/haze changes gradually, and in accordance, visibility changes gradually. It then is possible to construct a polynomials-fitted visibility variation curve and a piecewise stationary entropy visibility estimation function can be expressed as:

$$Vis = \alpha_n \times H_r^1 + \beta_n \times RH_r^2 + \gamma_n \times H_r^3 + \eta_n \quad (4)$$

where *Vis* is the final estimation value of foggy and hazy visibility. $\alpha_n$, $\beta_n$, $\gamma_n$ and $\eta_n$ are coefficients of the SPEV model with *n* ranging between 1 and 16. In formula (4), the power, intervals, interval ranges, and coefficients are obtained through big data training and manual calibration. It should be noted that there is a mirror relationship between the trend of Gaussian image entropy and real-world visibility in the time dimension. In practical applications, mirror flipping of the image entropy on the X-axis is required. The details of the algorithm proposed is shown in Table 1.

### B. Subjective assessment

In real-world applications, it is important to consider human visual perception. The visibility values, obtained by vision-based algorithms and optical instruments, will be used by staff of expressway management centers and by drivers. As such the visibility estimation values should be consistent with subjective assessments through human visual perception. We conducted a study based on the CIE visibility definition [26] with thirty-six subjects who performed observations during foggy and hazy visibility conditions. The results of SPEV were compared with those of subjective visibility assessments obtained through observations by the these subjects.

Thirty-six subjects were invited to observe all surveillance videos and the main task was to find the critical point of the visual range. The average value of the critical point coordinates observed by the thirty-six subjects was used as our measures for real-world observation visibility. Some physiological data for the human subjects, *e.g.*, visual acuity measures, are shown in Table 2. Five observational conditions were introduced: (1) 45 cm distance between eyes and computer screen; (2) Same room with the same lighting conditions; (3) Same computer monitor (FUJISTU, model B24T07, No. S26361-K1454-V160) set at the same brightness; (4) Five observations of each frame image and use of the median value; (5) The subjects were not allowed to engage in strenuous exercise or exhausting work prior to observations, they were asked to keep calm and be relaxed before the observations.

Using the critical point coordinates, fog/haze visibility could be calculated by formula (5) and (6) [9] below.

$$Vis' = \frac{\lambda}{v_v - v_h} \quad (5)$$

where:

$$\lambda = \frac{d_1 - d_2}{\frac{1}{v_1 - v_h} - \frac{1}{v_2 - v_h}} \quad (6)$$

here, $v_v$ is the critical point coordinate of the visual range, $d_1$ and $d_2$ are the two points on the expressway pavement, and $v_1$ and $v_2$ are their corresponding image coordinates. A problem discussed in the literature [9] is finding $v_v$ by machine automatically. However, the job is easy for humans. Therefore, we had human subjects mark $v_v$ in the foggy and hazy videos.

To illustrate the effectiveness of SPEV, a relative error measure was adopted for evaluating the estimation results:

$$APE = \frac{Vis - Vis'}{Vis} \times 100\% \quad (7)$$

where *Vis* denotes the estimation value of foggy- and hazy-day visibility, *Vis'* is the subjective assessment value.

**TABLE 1. SPEV algorithm**

| |
|---|
| Algorithm: SPEV algorithm |
| **Input:** Surveillance video, *1120 min × 60 s/min × 30 frame/s = 2,016,000* frames |
| **Output:** SPEV model, visibility (*Vis*) |
| **Step:** |
| 1. Surveillance video preprocessing |
|     (1) frame extraction; |
|     (2) de-noise; |
|     (3) ROI extraction: the sunny day images of the six surveillance points are used for extracting the expressway pavement, the ROI |
| 2. Compute the image entropy for the sunny day images |
|     (1) Compute the image entropy for all sunny day images (50 min) at the same surveillance point, based on formula (1)(2) and ROI extracted in step 1.3; |
|     (2) Remove the singular values; |
|     (3) Compute the mean value of image entropy. |
| 3. Compute foggy and hazy visibility; |
|     (1) Compute the image entropy for foggy and hazy images; |
|     (2) Compute the relative ratio of image entropy; |
|     (3) Mirror flip for the value in step 3.2; |
|     (4) Piecewise stationary function construction |
|         1) **Training set:** the coefficients $\alpha_n$, $\beta_n$, $\gamma_n$ and $\eta_n$ in formula (4) are obtained by training. |
|         2) **Testing set:** the data of a surveillance point (1, 2, or 4) is defined as the testing set, respectively. |
|             **Notes:** the data of one surveillance point are defined as the test set, the data of other surveillance points function as training sets. |
| 4. Optimize algorithm parameters by relative error validation. |

**TABLE 2. Anthropometric data (mean ± standard deviation) of human subjects.**

| Sample size | Sex | Age (year) | Height (cm) | Visual acuity (Eye chart) |
|---|---|---|---|---|
| 24 | Male | 23.67±5.16 | 177.13±7.37 | 1.275±0.26 |
| 12 | Female | 25.58±5.52 | 161.42±3.06 | 1.31±0.25 |

**TABLE 3. Parameter values of subjective assessment (Six surveillance points)**

| $v_{1\_15}$ | $v_{1\_9}$ | $v_2$ | $v_h$ | $\lambda_{15}$ | $\lambda_9$ |
|---|---|---|---|---|---|
| *544* | 577 | 650 | 289.45 | 12987.24 | 12781.79 |
| *733* | 791 | 919 | 311.41 | 20657.82 | 20488.87 |
| *805* | 885 | 1076 | 334.20 | 19330.81 | 19252.82 |
| *714* | 762 | 865 | 340.47 | 19463.06 | 19319.91 |
| *804* | 884 | 1076 | 335.42 | 19137.11 | 9043.72 |
| *829* | 880 | 992 | 455.48 | 18441.76 | 18302.38 |

**TABLE 4. Surveillance point Information**

| Surveillance point No. | Stakes | District | Start and end time | duration | date | Remarks |
|---|---|---|---|---|---|---|
| 1 | K113+000 | Dasheng | 06:30-09:50 | 200 min | 2016.4.14 | Test set |
| 2 | K148+150 | Haimen | 06:00-09:50 | 230 min | 2016.4.14 | Test set |
| 3 | K159+950 | Haimen | 06:00-09:50 | 230 min | 2016.4.14 | |
| 4 | K106+980 | Dasheng | 06:00-09:00 | 180 min | 2016.4.14 | |
| 5 | K159+950 | Haimen | 06:00-08:30 | 150 min | 2016.4.13 | Test set |
| 6 | K208+027 | Chenqiao | 06:00-08:10 | 130 min | 2016.3.15 | |

**TABLE 5. Piecewise stationary function coefficients**

| n | $\alpha_n$ (Power=1) | $\beta_n$ (Power=2) | $\gamma_n$ (Power=3) | $\eta_n$ (Power=0) | Sub-intervals |
|---|---|---|---|---|---|
| *1* | -359589.99 | 17288.14 | 0 | 1869889.56 | [0, 50) |
| *2* | 883340.36 | -42326.19 | 0 | -4608733.65 | [50, 60) |
| *3* | 26300.98 | -1265.49 | 0 | -136586.32 | [60, 70) |
| *4* | -6151.37 | 295.69 | 0 | 32067.06 | [70, 80) |
| *5* | -4589.69 | 221.27 | 0 | 23884.22 | [80, 90) |
| *6* | -3828.24 | 184.62 | 0 | 19939.19 | [90, 100) |
| *7* | -1811378.64 | 174704.69 | -5616.47 | 6260153.82 | [100, 120) |
| *8* | -2513504.75 | 241571.93 | -7738.93 | 8717405.82 | [120, 140) |
| *9* | 4160187.44 | -399757.58 | 12804.13 | -1.44E7 | [140, 160) |
| *10* | -690609.32 | 66312.66 | -2122.30 | 2397423.63 | [160, 180) |
| *11* | 272078.56 | -25969.77 | 826.40 | -950131.19 | [180, 200) |
| *12* | 8489387.73 | -812191.25 | 25900.81 | -2.96E7 | [200, 250) |
| *13* | 7112976.24 | -677067.67 | 21482.31 | -2.49E7 | [250, 300] |
| *14* | 4330993.72 | -408874.03 | 12866.64 | -1.53E7 | [300, 350] |
| *15* | 2912483.95 | -271901.84 | 8459.97 | -1.04E7 | [350, 400] |
| *16* | 3.36E8 | -3.19E7 | 1007556.71 | -1.18E9 | [400, 600] |

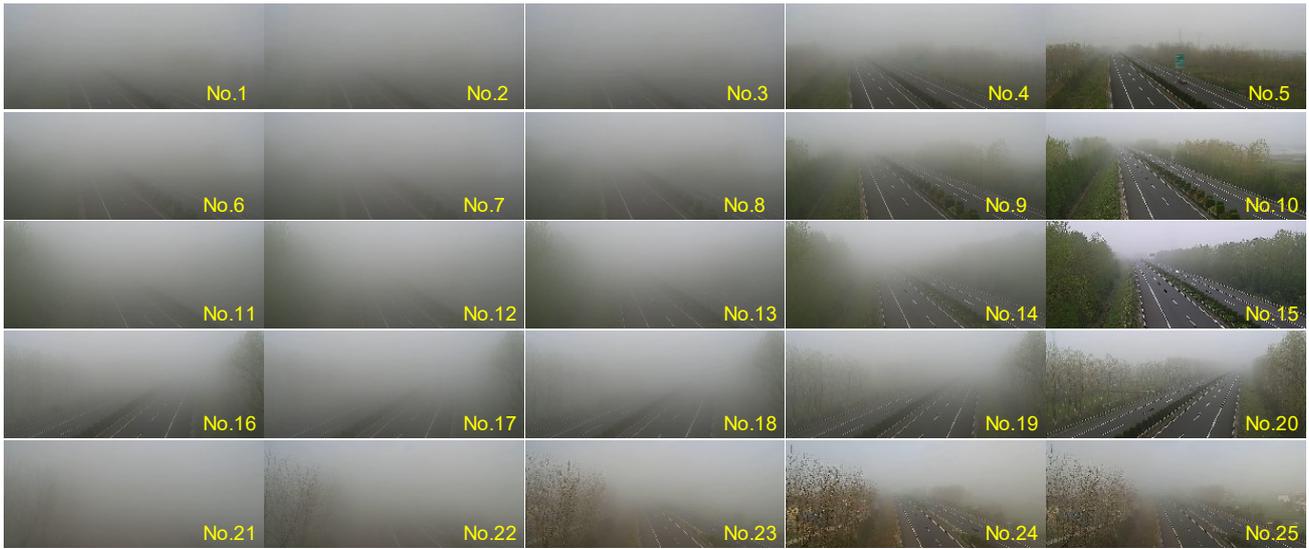

**FIGURE 1. Foggy and hazy images with different visibility**

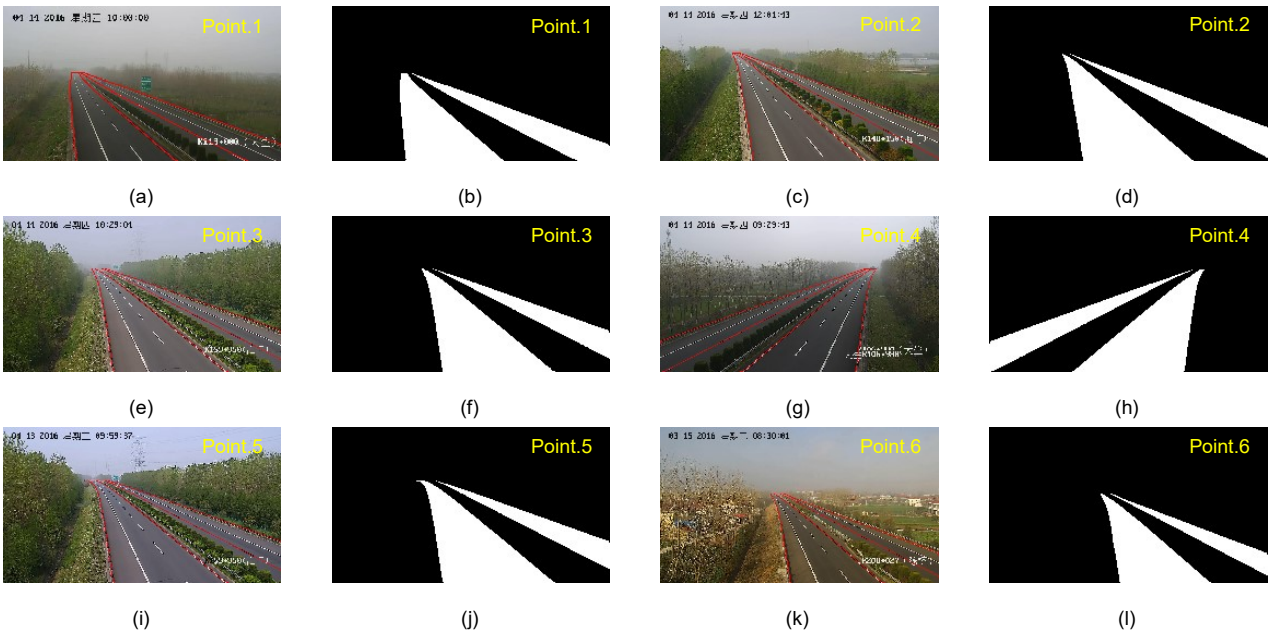

**FIGURE 2. Pavement extraction and overcoming estimation challenges of varying landscapes and brightness (A group of two sub-graphs, the pavement of the six different surveillance points are shown from Fig. 2-a to Fig. 2-i. ).**

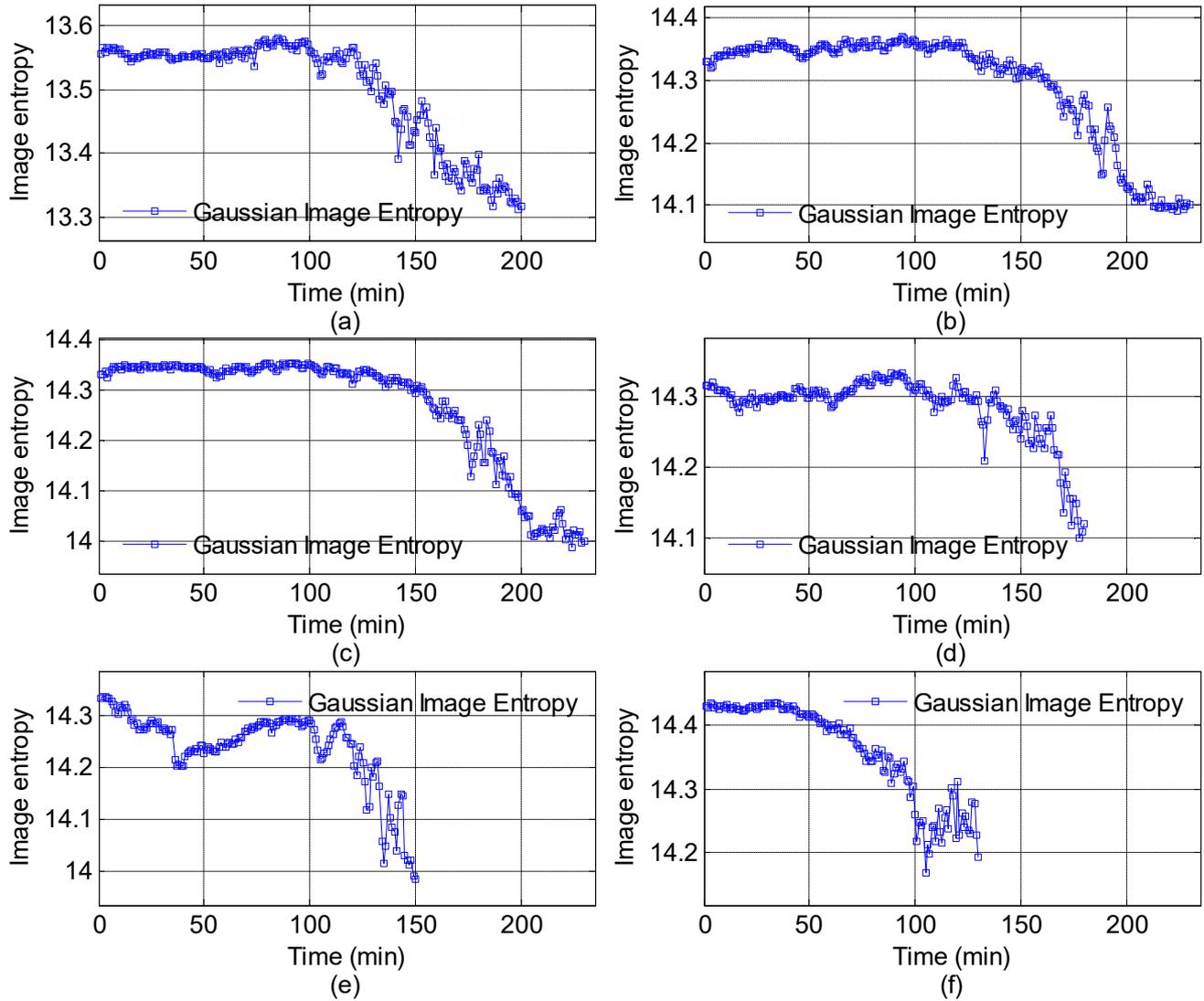

**FIGURE 3. Fogy and hazy image entropy (six surveillance points with different surveillance duration).**

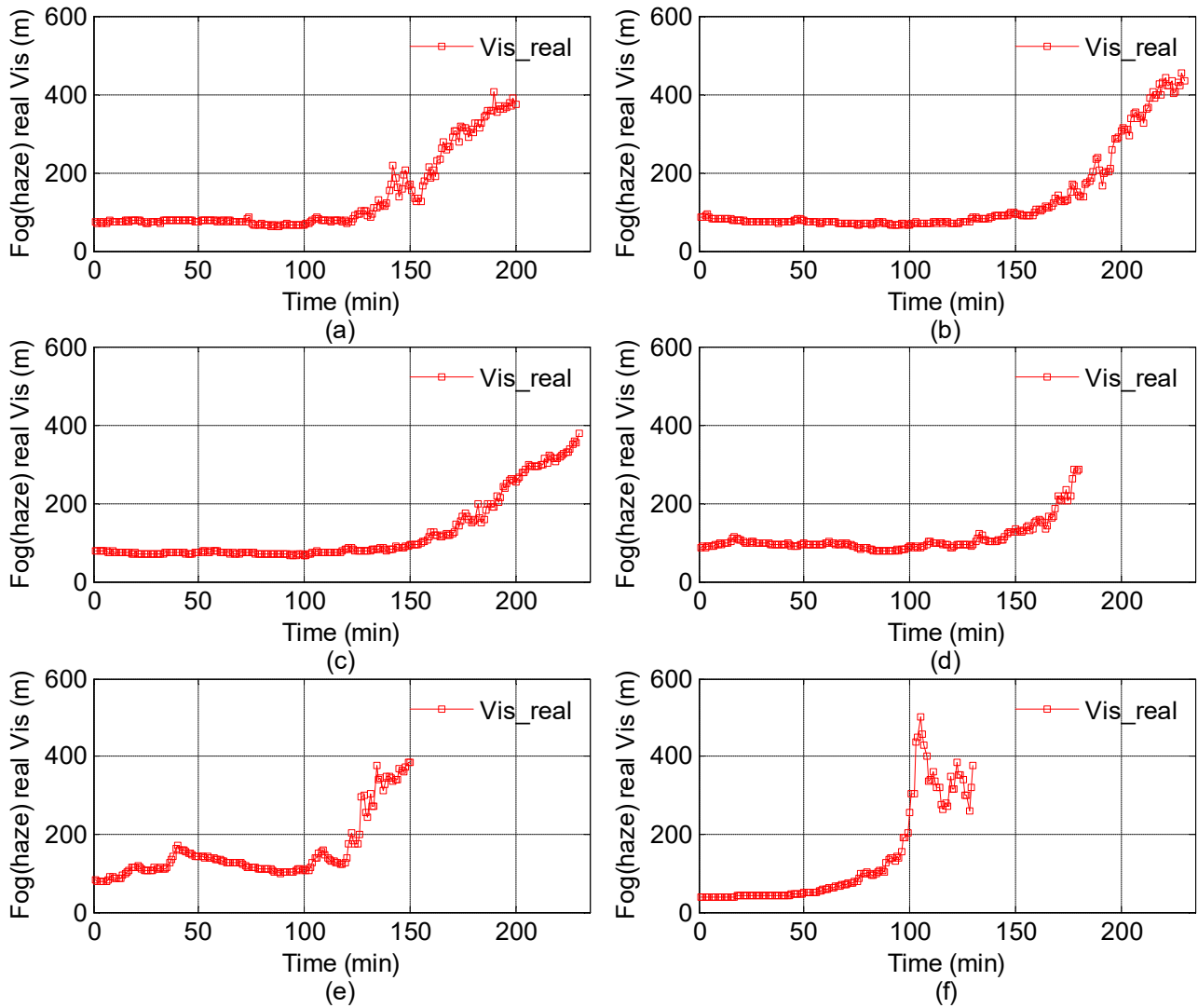

**FIGURE 4. Real visibility of foggy and hazy scenes (Six surveillance points with different surveillance duration).**

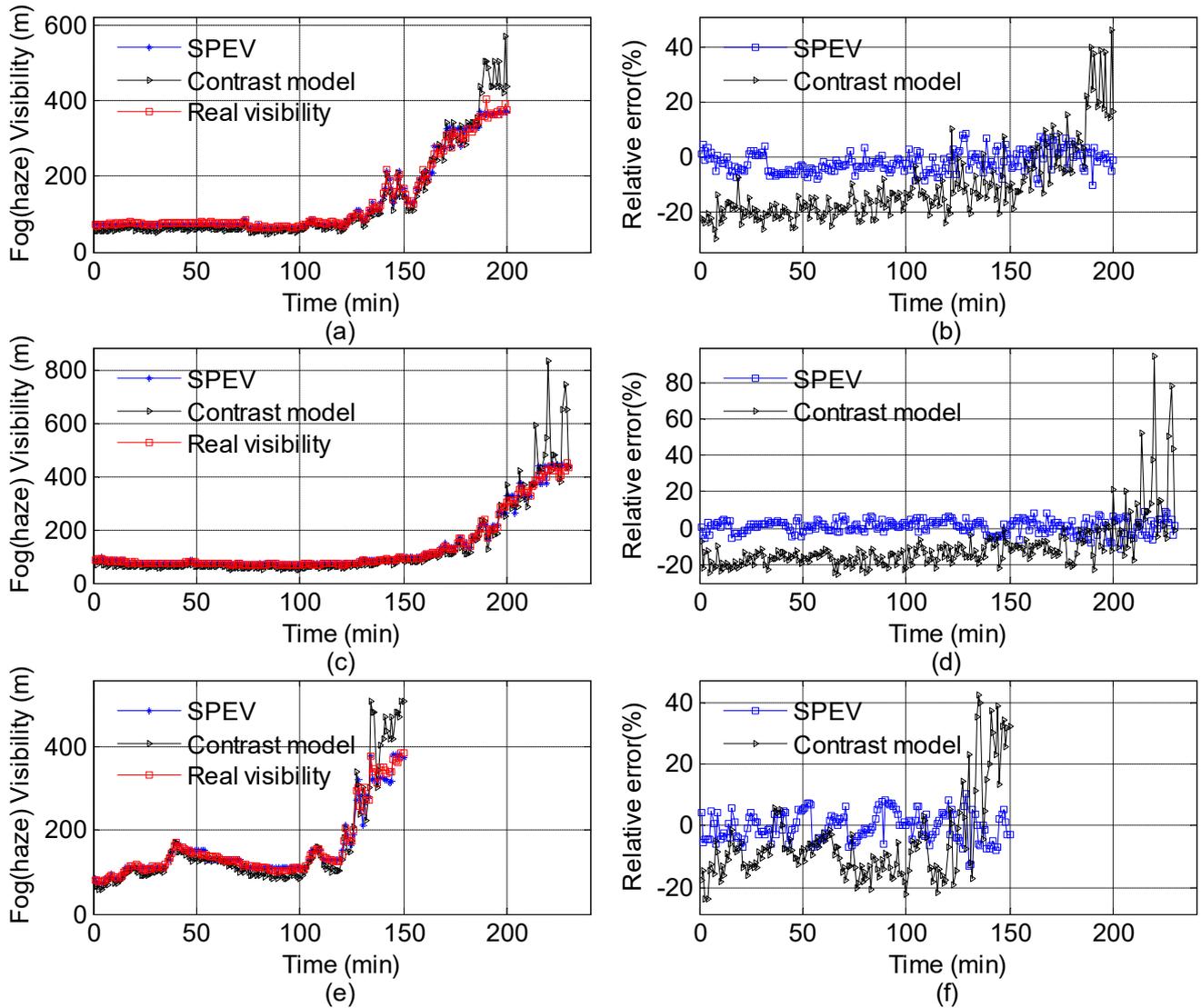

**FIGURE 5. Results of algorithm validation** (Data from a total of six surveillance points were used. The data of three surveillance points were assigned as test sets, respectively. For each test set, the data of the remaining surveillance points functioned as training sets. Fig. 5-a, c, e show visibility comparisons between visibility estimations of the SPEV algorithm, visibility estimations of the contrast model, and those of human observations. Fig 5-b, d, f show corresponding relative errors.).

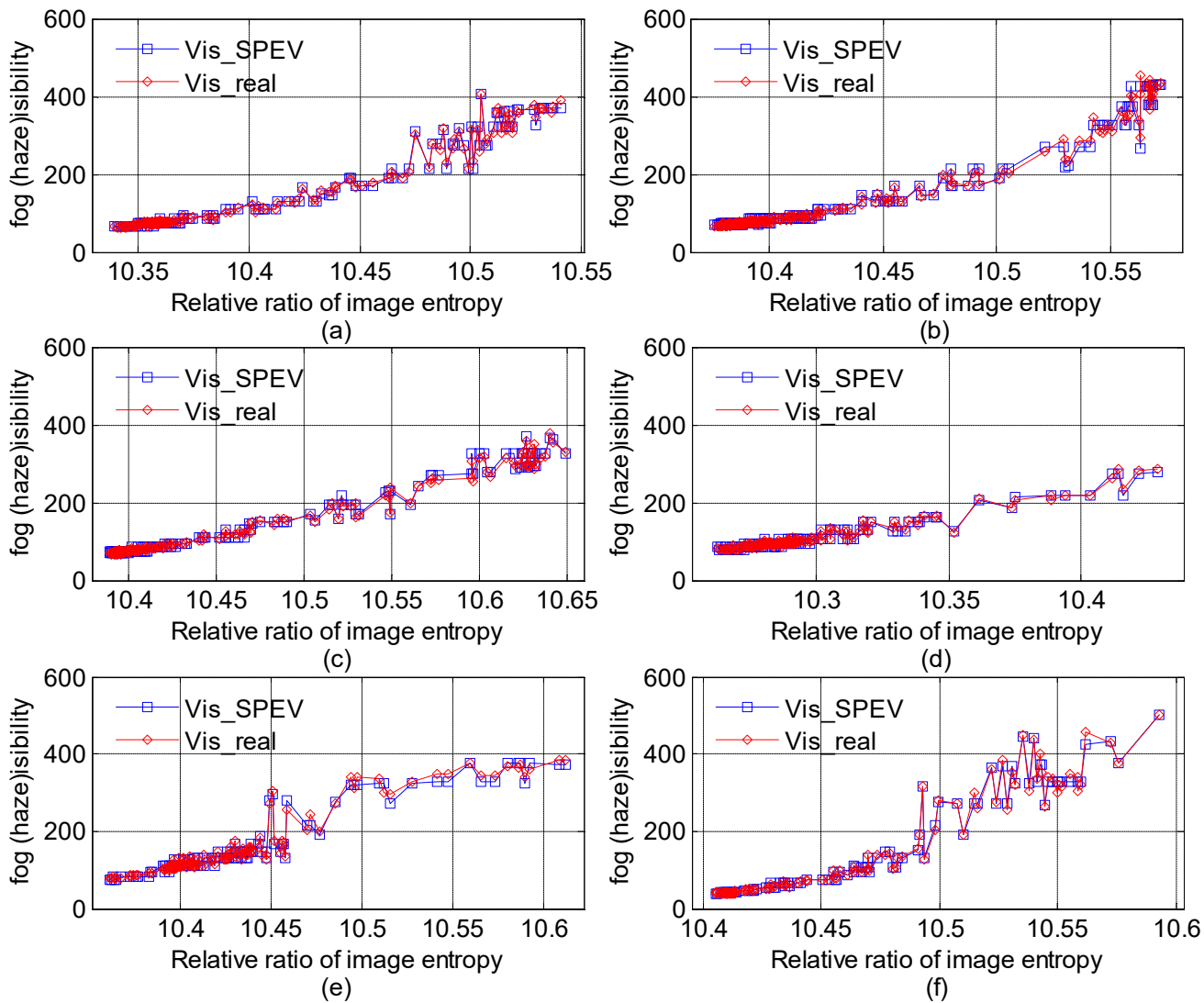

**FIGURE 6.** Entropy-visibility plot (six subgraphs correspond to the six surveillance points.).

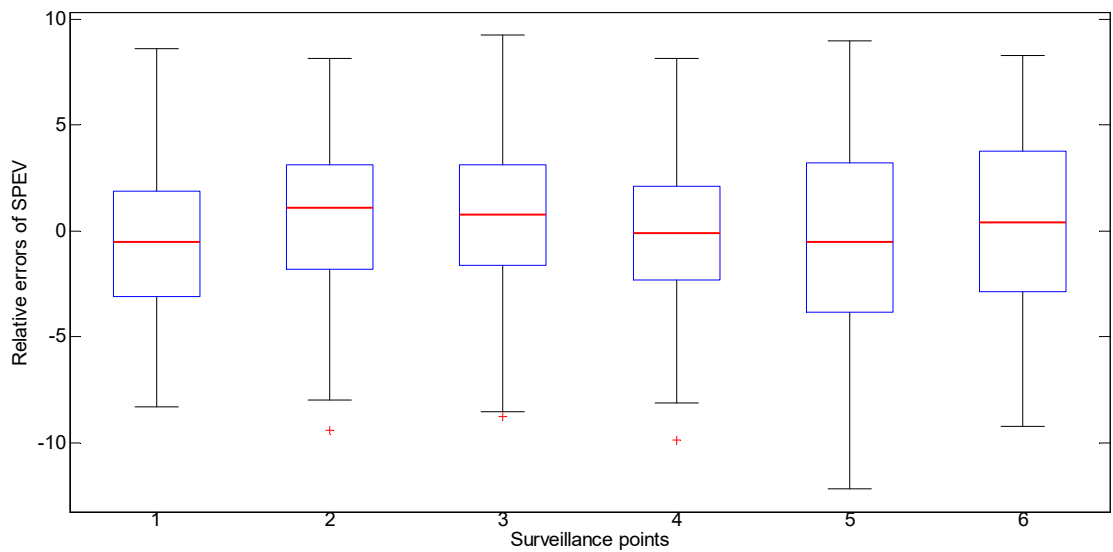

**FIGURE 7.** Relative error of SPEV algorithm (The six boxes denote the six surveillance points, respectively).

## IV. RESULTS

To validate SPEV as proposed in this paper, a big dataset was used for training and testing. The dataset, included foggy and hazy surveillance videos with different time durations, and was collected from the Tongqi expressway, China. The surveillance system there works around the clock.

We processed the subjective assessments for six surveillance points first. The assessment process is discussed in Section III. Some parameters used in SPEV are shown in Table 3 and explained as follows: (1) The subjects were required to assess all images 5 times and thus 5 values of $v_v$ could be obtained for each image and the median value was then adopted. (2) There are two options for the distances between $d_1$ and $d_2$ ($d_1$-$d_2$). The options are 15 *m* and 9 *m*. The lengths and intervals of the dotted lane line are 9 *m* and 6 *m* in China, respectively. The starting point and end point of the dotted lane line were extracted to calculate parameter $\lambda_{15}$ and $\lambda_9$. (3) Based on $\lambda_{15}$ and $\lambda_9$, two subjective visibility assessments for each image could be computed, and the mean value used as real visibility.

The hardware, used for presenting images to participants in our study was an X64-based dual core PC with 32 GB RAM, and an NVIDIA Geforce GTX 980 graphics card.

As shown in Table 4, six surveillance points of the expressway were selected for video collection. The resolution of the images were 1920 × 1080. A total of 2,016,000 frames (1120 min × 60 s/min × 30 frame/s) were used. The fog dissipated gradually from morning to noon, and the real visibility range of these data is [0*m*, 600*m*]. 25 image samples of foggy and hazy images extracted from the surveillance videos are shown in Fig. 1.

The background between different surveillance points varies considerably. The background includes trees, villages, farmland, *etc*. These landscape differences affect the calculation of image entropy. As shown in Fig. 2, the ROIs (stretches of expressway pavement) of six surveillance points, were extracted.

For each surveillance point, one clear-day image was selected to extract the ROI. Hough line detection was used in this process. Based on the ROI, the image entropy of all frames could be computed.

As shown in Fig. 3 and Fig. 4, there is a mirror trend relationship between image entropy and real visibility of fog and haze. When fog and haze visibility increases, the corresponding image entropy decreases. In the trend plot of the 6[th] surveillance point (Fig. 3-f, Fig. 4-f), the fluctuation of visibility is very close to that of image entropy, especially in the duration of [100 *min*, 150 *min*]. This shows that Gaussian image entropy can represent variation of foggy and hazy visibility.

As shown in Table 4, the data from surveillance points 1-2, and 5 were used for test sets. When data from one surveillance point was used as a test set, data from the others functioned as training sets. The test results are shown in Fig. 5. Fig. 5-a, c and Fig. 5-e show comparisons between visibility estimated by SPEV, visibility estimated by the contrast model, and real visibility estimated by humans. The corresponding estimation errors are shown in figures 5-b, d, and Fig. 5-f. A total of 580 images (200, 230, 150) were tested as shown in fig. 5. Only 5 relative estimation errors (SPEV) are greater than 10%. The relative errors (SPEV) are -10.26%, -11.48%, 10.51%, -12.89%, -12.42%, respectively. However, for the contrast model, 417 relative errors are more than 10%, and 113 relative errors are more than 20%. If we remove two singular values, the relative error of the contrast method is between -29.6277% and 52.5099%.

As a data-driven method, more data can make the algorithm model more robust. All 2,016,000 frames were used to obtain the coefficients of SPEV. The coefficients are shown in Table 5. Fig. 6 shows the scatter plots between estimated visibility and relative ratio of image entropy. The real visibility values were plotted for comparison. Fig. 7 shows the relative error of SPEV.

## V. DISCUSSIONS

The aim of SPEV is to solve the problem of practically useful visibility estimation. The visibility range of the dataset is [0 *m*, 600 *m*], as such the visibility range that the algorithm proposed can process is [0 *m*, 600 *m*]. Someone might object that it is a weak handling capacity. Based on the expressway management regulations, [0 *m*, 200 *m*] is the visibility range required to take precautionary measures, such as traffic flow control, speed limitation, closing the expressway, *etc*, but [200 *m*, 500 *m*] needs also be considered. The expressway is the most dangerous when the visibility is less than 200 *m*. Therefore, SPEV is well suited for visibility estimation on the expressway. In addition, to what we have considered in SPEV, atmospheric visibility offers additional complexities. It is not realistic to deal with all situations with one algorithm. In this paper, we just focus on the expressway situation.

In Fig. 3, we can see that the image entropy range is [13.3, 13.6]. However, the image entropy range of the other five surveillance points is [13.9, 14.6]. This difference will impact the adaptability of SPEV, if we apply the image entropy value to the equation (4) directly. As mentioned above, a relative ratio of image entropy is adopted in this paper and it is shown in Fig. 6, the value ranges of the X-axis is [10.3, 10.6], [10.3, 10.6], [10.3, 10.7], [10.2, 10.5], [10.3, 10.7], [10.4, 10.6], respectively. The difference is small and all relative ratios of image entropy are between 10 and 11. This shows that the ROI extraction (stretch of expressway pavement) and relative ratio of image entropy analysis were useful for visibility estimation.

According to [29-30], when the visibility is less than 2000 *m*, the error should be less than 10%. Figures 5 and 7, are based on a total of 580 and 1120 images, respectively. The number of relative errors more than 10% is 5 and 1 respectively. We consider fog and haze as a piecewise stationary signal. The results show that this assumption can be useful for visibility estimation.

The distances between surveillance points 4, 1, 2, and 3 are 6.02 *km*, 35.15 *km*, 11.8 *km* respectively. However, it is shown in Fig. 4-a, b, c, and Fig. 4-d that the visibility was different in the different places during the same time intervals. So the distribution of fog and haze on the expressway is not uniform over large distances. It is therefore necessary to perform visibility estimations of the expressway at each surveillance point.

As shown in Fig. 2, and Fig. 3, the variation trend of the entropy varies with the visibility changes. However, it should be noted that the histogram entropy is not suited for visibility estimation for long durations. There are many marked fluctuations in the curves of histogram entropy when the fog and haze visibility increase gradually (see appendix). This is the reason why we adopted Gaussian entropy analysis in this study.

## VI. CONCLUSIONS

SPEV, an improved vision-based visibility estimation algorithm is proposed in this paper. SPEV is based on image entropy and piecewise stationary time series analysis. The aims with SPEV is to overcome drawbacks of traditional vision-based methods for fog and haze visibility estimation, and reduce the number of expressway traffic accidents caused by fog and haze. There is one group of surveillance cameras every 5 *km* or every 1 *km*. SPEV can be integrated into the current intelligent transportation system to obtain a visibility map of the entire expressway. The conclusions of this paper can be presented as follows.

(1) It is the first time that Gaussian image entropy is used for estimating fog and haze visibility on an expressway.
(2) The SPEV algorithm proposed in this paper is a data-driven model. A big dataset (2,016,000 frames) was used for algorithm validation. The data was collected from the real world.
(3) The relative errors were encouraging, 99.14% of the relative errors were less than 10%. The utilization of ROI, relative ratio of image entropy, and the piecewise stationary time series analysis contributed to the encouraging results.
(4) Subjective assessments of fog and haze visibility were included. 36 subjects participated in the subjective evaluation in this study.

As to the limitation, more data are required to train SPEV. Deep learning techniques could also be used to construct the relationship between fog/haze images and real-world visibility. It would to make a comparison between the results of this study and an alternative approach based on deep learning. This will be our future work.

## ACKNOWLEDGEMENT

We would like to express our gratitude to Henrik Artman and Daniel Pargman. Thanks for their comments to this paper and discussion with us.

**APPENDIX**

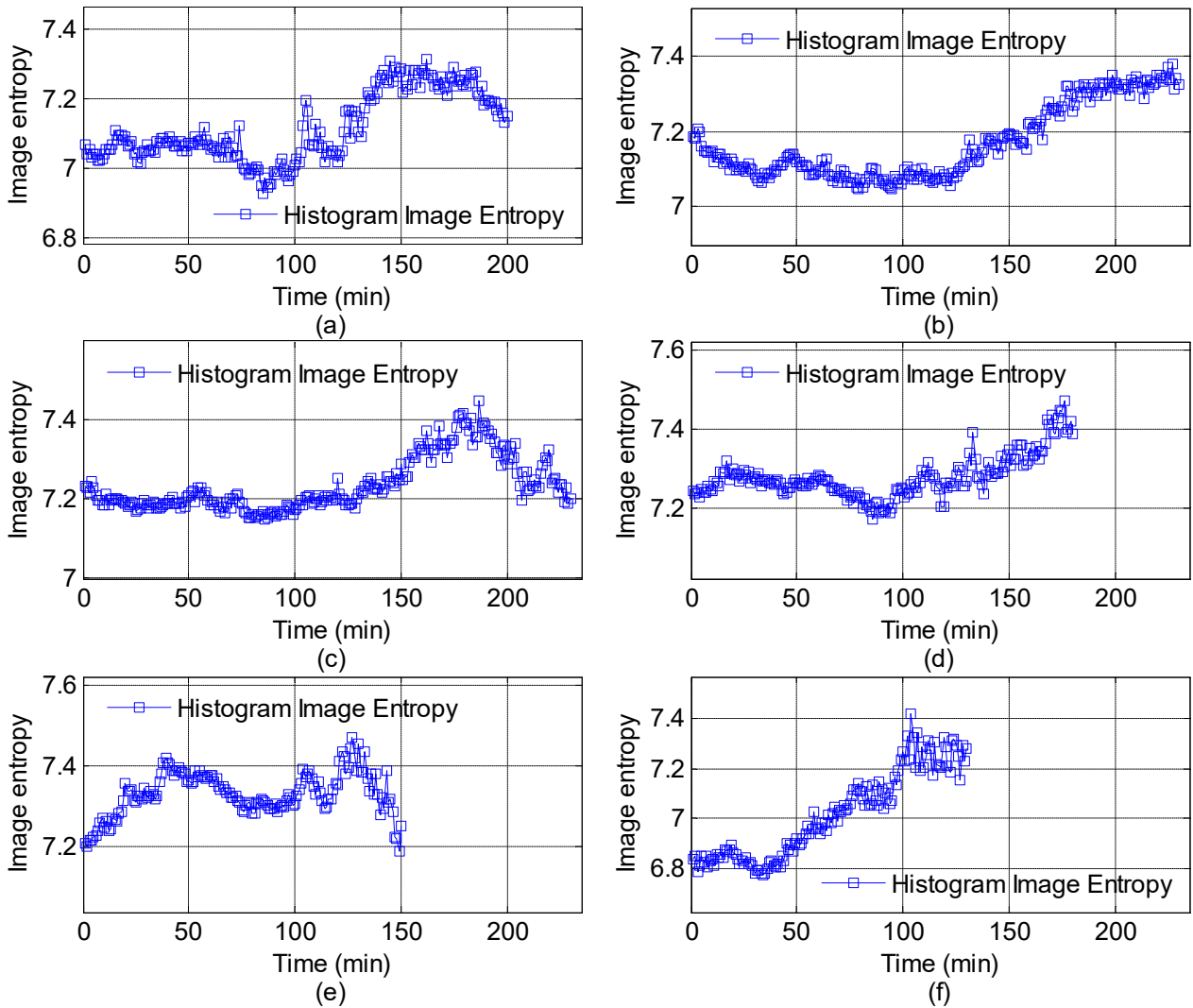

FIGURE 8. Histogram image entropy of fog and haze images (Notes: The variation trend, in detail, is different with the real visibility shown in Fig. 4. As such we adopt the Gaussian image entropy in this paper.).